\documentclass[sigconf]{acmart}
\usepackage{makecell}
\usepackage{multirow} 
\usepackage{multicol} 
\usepackage{arydshln}
\AtBeginDocument{%
  \providecommand\BibTeX{{%
    \normalfont B\kern-0.5em{\scshape i\kern-0.25em b}\kern-0.8em\TeX}}}

\copyrightyear{2022}
\acmYear{2022}
\setcopyright{acmcopyright}
\acmConference[MM '22] {Proceedings of the 30th ACM International Conference on Multimedia }{October 10--14, 2022}{Lisbon, Portugal.}
\acmBooktitle{Proceedings of the 30th ACM International Conference on Multimedia (MM '22), October 10--14, 2022, Lisbon, Portugal}
\acmPrice{15.00}
\acmISBN{978-1-4503-9203-7/22/10}
\acmDOI{10.1145/3503161.3548300}

\acmSubmissionID{2371}

\begin{document}

\title{Forcing the Whole Video as Background: An Adversarial Learning Strategy for Weakly Temporal Action Localization}

%

\author{Ziqiang Li}
\affiliation{%
	\institution{ Chongqing University}
	\city{}
	\country{}}
\email{ziqiangli@cqu.edu.cn}

\author{Yongxin Ge}
\authornote{Corresponding author.}
\affiliation{%
	\institution{Chongqing University}
	\city{}
	\country{}}
\email{yongxinge@cqu.edu.cn}

\author{Jiaruo Yu}
\affiliation{%
	\institution{ Chongqing University}
	\city{}
	\country{}}
\email{jiaruoyu@cqu.edu.cn}

\author{Zhongming Chen}
\affiliation{%
	\institution{Chongqing University}
	\city{}
	\country{}}
\email{ zhongmingcszmx@cqu.edu.cn}
%
%
%
%
%
%


\renewcommand{\shortauthors}{Li and Ge et al.}

\begin{abstract}
With video-level labels, weakly supervised temporal action localization (WTAL) applies a localization-by-classification paradigm to detect and classify the action in untrimmed videos. Due to the characteristic of classification, class-specific background snippets are inevitably mis-activated to improve the discriminability of the classifier in WTAL. To alleviate the disturbance of background, existing methods try to enlarge the discrepancy between action and background through modeling background snippets with pseudo-snippet-level annotations, which largely rely on artificial hypotheticals. Distinct from the previous works, we present an adversarial learning strategy to break the limitation of mining pseudo background snippets. Concretely, the background classification loss forces the whole video to be regarded as the background by a background gradient reinforcement strategy, confusing the recognition model. Reversely, the foreground(action) loss guides the model to focus on action snippets under such conditions. As a result, competition between the two classification losses drives the model to boost its ability for action modeling. Simultaneously, a novel temporal enhancement network is designed to facilitate the model to construct temporal relation of affinity snippets based on the proposed strategy, for further improving the performance of action localization. Finally, extensive experiments conducted on THUMOS14 and ActivityNet1.2 demonstrate the effectiveness of the proposed method.

\end{abstract}





\maketitle

\begin{figure}[t]
	\centering
	\includegraphics[scale=0.4]{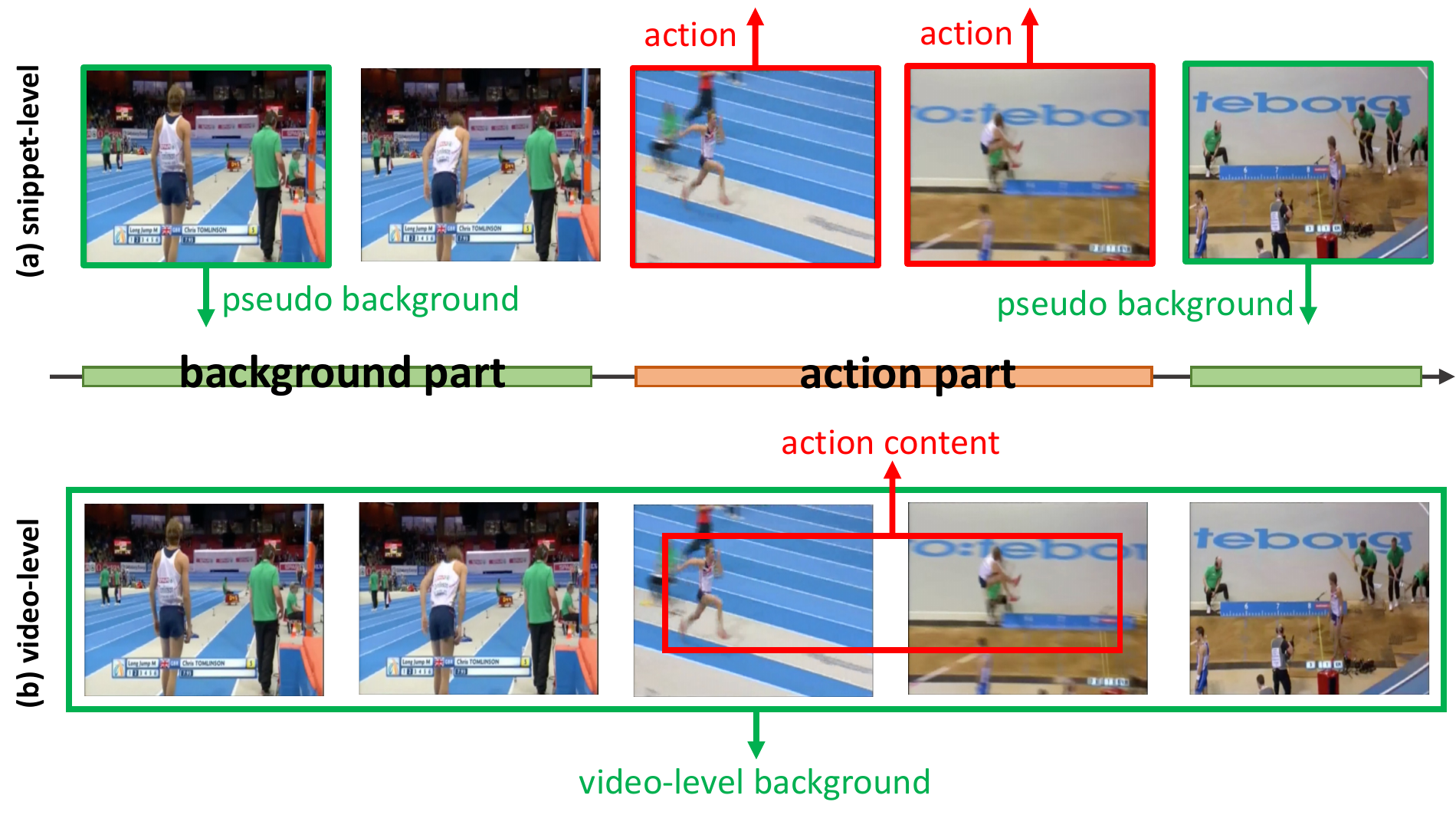}
	\caption{For long-jump instance, the background of the track field and sandpit appears in the whole video. (a) Previous works reply on artificial hypothetical to annotate pseudo background snippets. (b) An adversarial learning strategy is introduced that forces the whole video tend to the background. Under such conditions, the model has to boost its ability of action modeling for foreground recognition.}
	\label{fig_1}
\end{figure}

\section{Introduction}

Temporal action localization, which aims to localize and classify actions in untrimmed videos, has attracted great interest in the computer vision community due to its potential application in various fields~\cite{summarization, surveillance, highlight_detection}. In this task, fully supervised methods, owning snippet-level labels, have triggered remarkable progress. However, with the rapid growth of videos on various websites, annotating the precise action instance boundary in untrimmed videos is expensive and time-consuming. Consequently, weakly supervised temporal action localization (WTAL) has drawn a lot of attention recently, where cheaper video-level labels are introduced.

Different from the fully supervised counterpart, most existing WTAL methods adopt a ‘localization by classification’ paradigm~\cite{w2_cola}. However, without snippet-wise labels, the classifier is prone to focus on salient features to boost its discriminability for classification~\cite{discri_relational}. Therefore, the background snippets with class-specific content in untrimmed videos would be misrecognized as foreground(action) by the classifier in WTAL methods, hindering the separation of background and actions~\cite{w3_gam}. Simultaneously, the capacity for action modeling of the classifier is weakened due to the preference for class-specific background snippets.

To alleviate the disturbance of the background, some classical approaches are proposed. Lee et al.~\cite{w14_um} assume that background snippets could be dynamic and inconsistent, and model the background uncertainty by suppressing the feature magnitudes of pseudo background snippets with bottom-k scores. To further distinguish class-specific background (context) and action, ACSNet~\cite{w1_acsnet} introduces auxiliary categories for representing the class-specific background. Furthermore, CoLA~\cite{w2_cola} utilizes the fruitful temporal context relation to mine pseudo hard background and easy action snippets and conducts contrastive learning on them for better action modeling. All of the above works attend to explicitly model background snippets to further boost the discrepancy between background and actions, leading to precise localization. Nevertheless, these approaches largely rely on artificial hypotheticals for class-specific (hard) background snippets annotating, and this priority inevitably marks some snippets of action into pseudo annotations.

Let’s summarize the core challenges on WTAL: 1) preference for class-specific background blocks the robust capacity from action modeling, and 2) fully filtering out background snippets is impractical due to the absence of snippet-level labels. As mentioned above, the previous methods attempt to model background content by pseudo label annotating at the snippet level, as shown in Fig.~\ref{fig_1}(a). However, the representation of background occurs in the whole untrimmed videos. To illustrate it, we take the long jump in Fig.~\ref{fig_1} as an example. Track-field and sandpit, as background content, appear both in the background and action snippets along the temporal dimension. Naturally, each snippet in untrimmed videos owns the tendency to be rated as background. 

Inspired by the above analysis, distinct from the previous works that model background by pseudo-class-specific background annotating, we design a strategy to explicitly model action instances under the circumstance that the whole video is regarded as background. Concretely, motivated by GAN~\cite{GANgoodfellow2014}, we adopt an adversarial learning strategy based on the foreground and the background classification losses. The background loss confuses the model by forcing the whole untrimmed video to tend to be background, while the foreground loss drives the model to focus on the action snippets. And competition between the two losses boosts the ability for action modeling to distinguish the foreground from the video-level background, reducing the mis-recognization caused by the class-specific background. Simultaneously, to complement the adversarial learning strategy, we also explicitly construct the temporal relation of snippet sequences to further improve the model for localizing complete action. Accordingly, achieving the adversarial learning procedure and modeling the temporal relation of snippet sequence are the core tasks in our work.  

In this paper,  we implement the proposed method for precise localization by fulfilling the following tasks. \textbf{T1: Steer the whole video toward the background.} We propose a new background gradient enhancement strategy (BGES) that increases the gradient of each snippet towards to background in the backward procedure by modifying the background classification loss. In this manner, the whole video attends to be regarded as the background in training procedure for raising the difficulty of classification and the model correspondingly has to self-boost its capacity of perceiving action, achieving the procedure of adversarial learning. \textbf{T2: Further boosting the ability for action modeling.} We also present a novel temporal enhancement network (TEN) to model the temporal relation of affinity snippets, which consists of the base branch (BB), and the temporal continuity branch (TCB). Functionally, BB is the baseline of our method and TCB is an auxiliary one for it. Simply, the inputs of BB and TCB are the sequence pairs of affinity snippets. With the interaction of BB and TCB, the temporal continuity of the neighboring snippets is obtained by the model, which promotes the performance of the completeness modeling. Our full method is the combination of BGES and TEN. Extensive experiments conducted on THUMOS14 and ActivityNet demonstrate the effectiveness of the proposed method.

To sum up, the main contributions of our works are as follows: (1) Pioneeringly, we design the adversarial learning paradigm for WTAL to drive model focus on action under the encompassment of background. (2) A simple yet efficient strategy is introduced to guide the whole video to tend to be the background in the gradient backward process. (3) A novel temporal enhancement network (TEN) is proposed to explicitly perceive temporal relationships, aiming to model action completeness.

\section{Related Work}

\noindent\textbf{Fully Supervised Temporal Action Localization.}
The goal of temporal action localization (TAL) is to predict the precise start time and end time of action instances in an untrimmed video and recognize the corresponding action category. The fully supervised paradigm of TAL, relying on snippet-level annotations, has drawn lots of interest recentlly~\cite{full10_graph,full8_zhao}, and it can be divided into two types. The methods~\cite{full7_tcn,full12_fast} of the first type generally adopt a two-stage pipeline that includes proposal generation and instance classification. These two-stage methods mainly aim to enhance the localization performance, such as improving temporal proposals searching~\cite{full2_bsn,full3_bmn} and proposing robust classifiers~\cite{full1_ssn}. Compared with the former type, the second type of works~\cite{full16_long2019gaussian,full6} is more flexible and efficient, which applies a one-stage mode to generate proposals and classify categories simultaneously. Although the above methods make remarkable results, they are highly laborious-costing for the manual annotating timestamp of action instances.

\noindent\textbf{Weakly Supervised Temporal Action Localization.}
Distinct from the fully supervised TAL, the weakly supervised methods~\cite{w4_completeness,w5_bm} mostly adopt video-level labels in training procedures, reducing the labor-intensive annotating cost. STPN~\cite{w13_stpn} identifies a set of key segments with an attention mechanism and aggregates them by adaptive temporal pooling. To strengthen the confidence of foreground scores, Yang et al.~\cite{w8_uncertainty} present an uncertainty-guided collaborative training strategy to mitigate the noise of pseudo action snippets and FAC-Net~\cite{w6_foreground} takes the bilateral relations of foreground and action into consideration and guarantees the consistency of foreground and action. Moreover, some methods~\cite{w12_liu,w16_completeness_ban} are proposed to explicitly model action instances. 3C-Net~\cite{w19_c3} introduces extensive information by counting the number of action instances, and TSCN~\cite{w11_tscn} utilizes the consistency of RGB and optical flow modality to focus on action. Furthermore, Luo et al.~\cite{w10_aum} take into account the complexity of action and propose an Action Unit Memory Network. Additionally, some attempts~\cite{w3_gam,w17_backgroundsuppression}  are present to mitigate the background interference. ACSNet~\cite{w1_acsnet} introduces an extensive category to explicitly model background, and CoLA~\cite{w2_cola} proposes a snippet contrastive loss to refine hard background samples. Lee et al.~\cite{w14_um} assume that background snippets could be dynamic and inconsistent and suppress the feature magnitudes of pseudo background snippets with bottom-k scores.   

\noindent\textbf{Adversarial Learning.}
Following the early work GAN~\cite{GANgoodfellow2014}, the idea of adversarial learning is widely applied in various fields~\cite{G1_image,G5_adversarial}. For example, Isola et al.~\cite{G1_image} propose conditional adversarial networks (CGAN) as a general-purpose solution for image-to-image translation. Compared with CGAN, Circle-GAN~\cite{G2_unpaired} is proposed for translation learning in the absence of paired samples. In addition, ADDA~\cite{G5_adversarial} adopts a novel unified framework for adversarial domain adaptation based on adversarial learning. Furthermore, Ganin et al.~\cite{G4_unsupervised} present a gradient reversal layer (GRL) to minimize the discrepancy between feature distributions from the target and source domains. In our work, we force the whole video to be regarded as background in the gradient backpropagate procedure, driving the recognition model to focus on action semantic features by adversarial learning.

\begin{figure*}[t]
	\centering
	\includegraphics[scale=0.86]{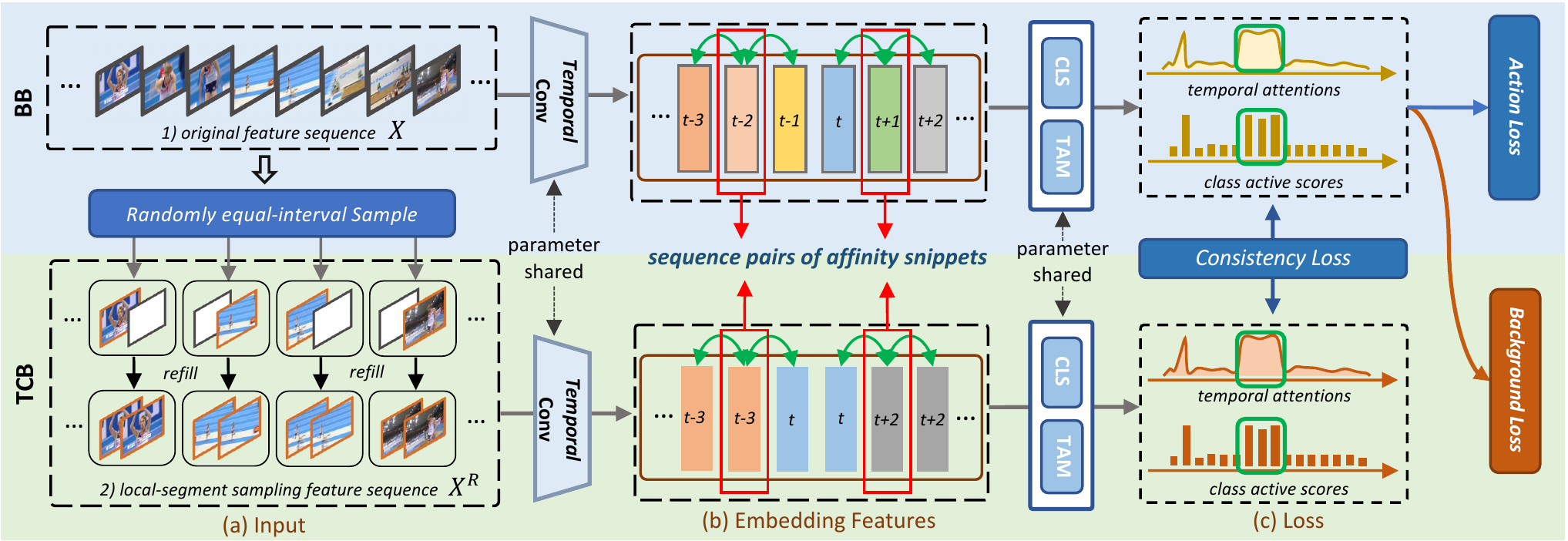}
	\caption{A pipeline of temporal enhancement network. (a) Given the original feature sequence $ X $, we randomly sample and refill them at equal intervals to get a new feature sequence $ X^R $. (b) And $ X $ and $ X^R $ are fed into the model to generate the embedding-feature sequence pairs of affinity snippets. (c) Then the feature sequence pairs are utilized to get the attention pairs and the CAS pairs, in which the related losses are conducted. }
	\label{fig_2}
\end{figure*}

\section{The Proposed Method}

\textbf{Problem Definition.}
Assume that we have $ N $ untrimmed videos $\{v_i\}_{i=1}^N $ with their corresponding video-level labels $\{\tilde{y}_i\}_{i=1}^N $, where  $\tilde{y}_i\in {\mathbb R}^C $ and $\tilde{y}_{i,c}$ means the presence/absence of $c$-th action classes. Following recent methods~\cite{w10_aum,w11_tscn}, for each video $v_i$, we divide it into non-overlapping snippets and feed them into a pre-trained I3D~\cite{i3dandk} to generate features as the input of our method. The features $ X^r\in {\mathbb R}^{T\times D}$ and $X^o\in {\mathbb R}^{T\times D} $ are extracted from the RGB and the optical flow modality snippets, where $ T $ and $ D $ indicate the number of snippets and dimension of features. During inference, we attempt to generate a set of action instances $\{c_i,q_i,s_i,e_i\}$, where $c_i$ and $q_i$ are the action class and confidence score, $s_i$ and $e_i$ represent the start time and end time of action instances correspondingly.

\textbf{Overview.}
We propose a novel temporal enhancement network (TEN) based on a background gradient enhancement strategy (BGES), which achieves the adversarial learning process to improve the model’s ability for action localization. BGES is applied in the backward procedure of background loss. In this manner, the background classification loss confuses the recognizing model by forcing the whole video to tend to be the background. On the contrary, the foreground classification loss guides the model to obtain action snippets under this condition. Consequently, competition between the two losses drives the model to self-boost the capacity for action modeling. Additionally, TEN is designed to explicitly enhance the temporal relation of affinity snippets for further improving the performance of the model in the adversarial learning procedure. For clarification, we first describe TEN and then introduce the BGES.



\subsection{Temporal enhancement network}
To model action completeness, Hide-and-seek~\cite{hideandseek} adopts a random hidden strategy on snippet sequences, and Liu et al.~\cite{w4_completeness} propose a multi-branch network where each branch is utilized to focus on different action content of one snippet.  However, all these works ignore the temporal properties of action instances. Consequently, the temporal enhancement network (TEN) is designed to promote the model to construct temporal relation of snippet sequences, which further boosts the capacity of the model for action modeling, leading to the localization completeness of action. As illustrated in Fig.~\ref{fig_2}, TEN is composed of two branches, namely, the base branch (BB) and the temporal continuity branch (TCB), and the two branches are built on the same parameter-shared model. 

\textbf{Base branch.} BB is the baseline of our method, and here we introduce it to describe the model which consists of three parts, namely the feature embedding module, classifier, and temporal attention mechanism.


\textit{Feature embedding.} Due to the features $X^r\in {\mathbb R}^{T\times D}$ and $X^o\in  {\mathbb R}^{T\times D}$ are respectively extracted from the RGB and the optical-flow modality snippets, the distributions along the channel dimension of them are inconsistent. To bridge the modality domain gap, we leverage the independent temporal convolutional layers $ {\mathrm\Phi}_r $ and $ {\mathrm\Phi}_o $ to embed $ X^r $ and $ X^o $, for learning a new set of features that can be applied in our specific task. Simply, this process can be defined as:
\begin{equation}
X_e^r={\mathrm\Phi}_r(X^r, w_e^r), \qquad   X_e^o={\mathrm\Phi}_o(X^o, w_e^o),
\label{eq_0}
\end{equation}
where $ X_e^r \in  {\mathbb R}^{T\times E}$ and $ X_e^o \in  {\mathbb R}^{T\times E}$ are the embedding features from the RGB and the optical-flow modalities, $ w_e^r $ and $ w_e^o $ are learnable parameters, and $ E $ represents the corresponding dimension of the feature. And $ {\mathrm\Phi}_r $ and $ {\mathrm\Phi}_o $ are implemented with single temporal convolutional layer following a ReLU function.

\textit{Classifier (CLS).} As a localization-by-classification paradigm, each snippet in the untrimmed video is predicted a class score for applying thresholding to localize action instances. Therefore, given the embed features $ X_e^r $ and $ X_e^o $, we utilize the classifiers to generate snippet-level class activation scores (CAS) $ y_r \in  {\mathbb R}^{T\times (C+1)}$ and $ y_o \in  {\mathbb R}^{T\times (C+1)}$ under the RGB and the optical-flow modalities. This procedure can be formulated as follows:
\begin{equation}
y_r={X_e^r}{w_{cls}^r}+b_{cls}^r, \qquad   y_o={X_e^o}{w_{cls}^o}+b_{cls}^o,
\label{eq_1}
\end{equation}
where $ w_{cls}^r\in  {\mathbb R}^{E\times (C+1)} $ and $ w_{cls}^o\in  {\mathbb R}^{E\times (C+1)} $ are the weights of the classifiers, $ b_{cls}^r $ and $ b_{cls}^o $ are the biases, and the $ C+1 $ represents the background class. Given the CASs of the RGB and the optical-flow modalities, we fuse them to get the  final CAS $ y $ of untrimmed videos, i.e. $ y=(y_r+y_o)*0.5 $.

\textit{Temporal attention mechanism (TAM).} Following the previous attempts~\cite{w3_gam,ham}, we also introduce TAM to recognize action snippets. And TAM is considered as a foreground classifier to produce the probability that each snippet belongs to the action. Similar to the classifiers, the embedded features $ X_e^r $ and $ X_e^o $ are leveraged to calculate class-agnostic temporal attentions $ a_r \in  {\mathbb R}^{T}$ and $ a_o \in  {\mathbb R}^{T}$ of the two modalities:
\begin{equation}
a_r=\sigma({X_e^r}{w_{att}^r}+b_{att}^r), \qquad   a_o=\sigma({X_e^o}{w_{att}^o}+b_{att}^o),
\label{eq_2}
\end{equation}
where $ w_{att}^r\in  {\mathbb R}^{E} $ and $ w_{att}^o\in  {\mathbb R}^{E} $ are the learnable parameters of temporal attention mechanisms, $ b_{att}^r $ and $ b_{att}^o $  are the corresponding biases, and $ \sigma(\cdot) $ represents the \textit{sigmoid} activation function. And the final class-agnostic attentions $ a $ is calculated by aggregating $ a_r $ and $ a_o $, i.e. $ a=(a_r+a_o)*0.5 $.

Intuitively, the class-agnostic attention and action-class scores should be consistent, i.e. given the $ i $-th snippet that belongs to the $c$-th class of action, the values of $ a_i $ and $ y_{c,i} $ are similar. Naturally, we fuse the CAS  $ y $ and class-agnostic attention $ a $ to get the video-level action class probability $P_{fg}\in \mathbb R^{C+1}$ and the corresponding background probability $P_{bg}\in \mathbb R^{C+1}$:
\begin{equation}
P_{fg}=\uptau(\frac{\sum_{i=1}^T{y_i\cdot a_i}}{N_f}), \qquad P_{bg}=\uptau(\frac{\sum_{i=1}^Ty_i\cdot(1-a_i)}{N_b}),
\label{eq_3}
\end{equation}
where $\uptau$ represents a \textit{SoftMax} function, $N_f=\sum_{i=1}^Ta_i$ and $N_b=\sum_{i=1}^T(1-a_i)$ are the normalized factors. Notably, $ a_i $ is the foreground score of the $ i $-th snippet and $ 1-a_i $ represents the background one.

Finally, given the video-level probabilities of the action and the background categories, the cross-entropy losses are conducted to optimize the model: 
\begin{equation}
{\mathcal L}_{fg} =-\sum_{c=1}^{C+1}\tilde{y}_c \cdot \mathrm {log}(P_{fg}^c), \qquad {\mathcal L}_{bg}= - \mathrm {log} (P_{bg}^{C+1}),
\label{eq_6}
\end{equation}
where $ c $ represents the action class, and $ \tilde{y} \in \mathbb R^{C+1}$ is the ground-truth label. $ {\mathcal L}_{fg} $ is the foreground classification loss to guide the model to focus on action snippets, and $ {\mathcal L}_{bg} $is the background classification one that makes the model distinguish the background from the action snippets.  

\textbf{Temporal continuity branch.} Due to the temporal continuity of the snippet sequence, affinity snippets tend to be consistent in content and category. Therefore, TCB is introduced as an auxiliary unit to interact with BB, aiming to constrain the temporal consistency of the contiguous snippets. Since TCB and BB are built on the same model, we simplify the description of TCB and adopt $ X $ to denote both RGB and optical flow features.

As shown in Fig.~\ref{fig_2}, given the extracted feature sequence $X=[X_1,X_2,\cdots,X_T]\in \mathbb R^{T\times D} $ of a video, we divide it into $ \frac{T}{k} $ segments by equal-interval and randomly select one snippet from each segment to get a sequence $X=[X_k,\cdots,X_{T-K}]\in \mathbb R^{\frac{T}{k}\times D} $. Then, we uniformly repeat the sampled snippet in each divided segment until to recover the original length, and obtain a new sequence  $ X^R\in \mathbb R^{T\times D}$. Similar to the sampling manner proposed in TSN~\cite{tsn},  this way still makes $ X^R $ retain the temporal characteristic of the original sequence $ X $, and $ X $ and $ X^R $ are conducted as the sequence pair of affinity snippets. Thereafter,  $ X^R $ is fed into the model to generate the corresponding action scores $ y^R\in \mathbb R^{T\times (C+1)} $ and temporal foreground attentions $ a^R\in \mathbb R^{T} $. Here, $ (a, a^R) $ and $ (y, y^R) $ are the score pairs of affinity snippets, and we conduct constraint losses on them to guide the model to learn the consistent information of the continuous snippets. For the temporal attention of BB and TCB, the $ L_1 $ normalized function is utilized to optimize them:
\begin{equation}
\mathcal L_{att}=\frac{1}{T}\sum_{i=1}^T{\left| a_i - G(a^R_i)\right| +\left| a^R_i - G(a_i)\right|},
\label{eq_11}
\end{equation}
where $ G(\cdot) $ represents the gaussian function among temporal dimensions, and it is utilized to smooth attention scores. Motivated by the success in knowledge distillation~\cite{klhinton}, a KL diverse loss is applied on the class activate scores to transfer  information of affinity snippets: 
\begin{equation}
\mathcal 
L_{KL}=\frac{1}{T}\sum_{i=1}^T\{KL(\uptau(y_i^{R})||\uptau(y_i))+KL(\uptau(y_i)||\uptau(y_i^{R}))\},
\label{eq_12}
\end{equation}
where $KL(p||q)$ is the KL divergence of $ p $ from $ q $. Consequently, by the multual learning procedure, the above constraint losses facilitate the model to learn the consistency of continuous snippets, improving the model for modeling action continuity. 

\textbf{Joint training.} The action and the background classification losses are solely conducted on BB, and we combine all the losses above to optimize our model:
\begin{equation}
\mathcal L_{all}=\mathcal L_{fg}+\lambda\cdot\mathcal L_{bg}+\beta\cdot (\mathcal L_{KL}+ \mathcal L_{att}),
\label{eq_13}
\end{equation}
where $\lambda$ and $\beta$  represent the hyper-parameters to adjust the balance among the losses.

\subsection{The background gradient enhancement strategy}
\begin{figure}[t]
	\centering
	\includegraphics[scale=0.6]{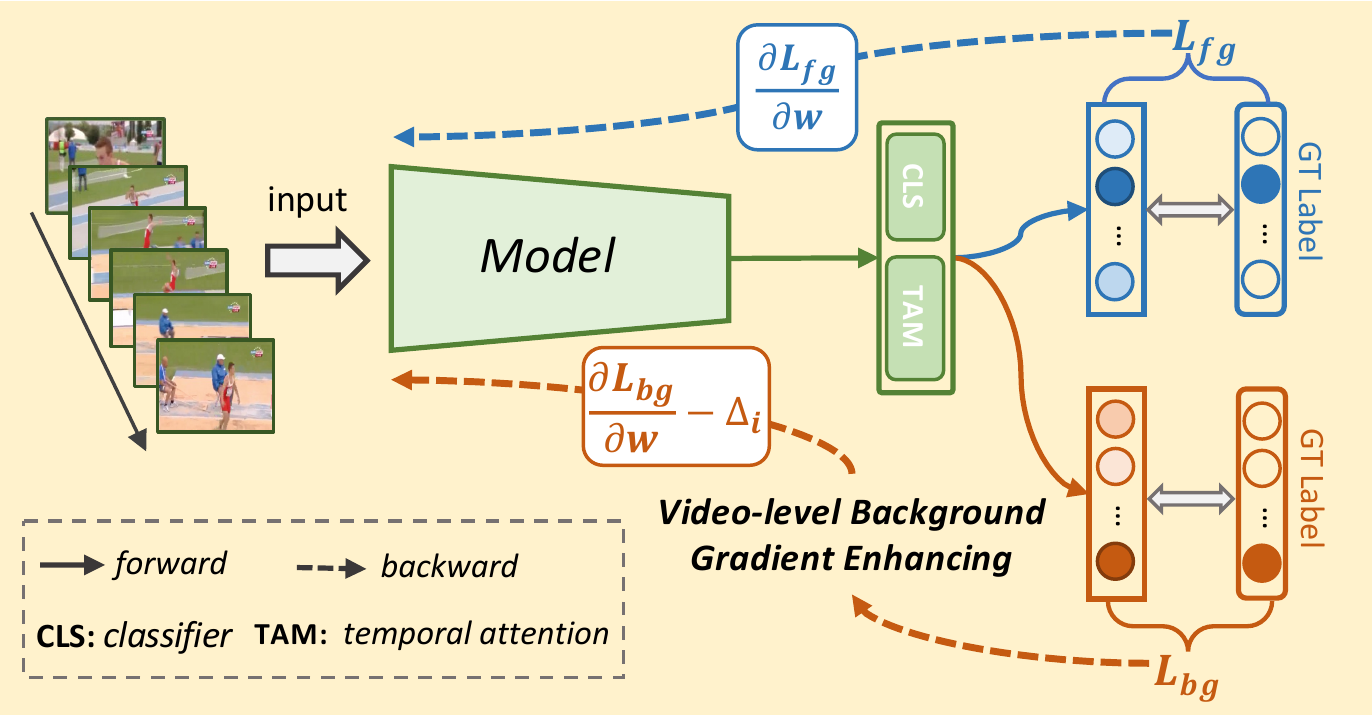}
	\caption{Illustration of the forward and the backward procedures of the model. And the background gradient enhancement strategy is applied to the background classification loss.}
	\label{fig_3}
\end{figure}

The class-specific background tends to co-occur with the action snippets~\cite{w3_gam}, hindering the removals of background snippets from localized action instances. Motivated by GAN, we aim to construct an adversarial learning procedure that the background classification loss $ {\mathcal L}_{bg} $  guides the model to regard the whole video as background in the backward procedure while the foreground classification losses $ {\mathcal L}_{fg} $  drive it to focus on action, as illustrated in Fig.~\ref{fig_3}. As a result, competition between $ {\mathcal L}_{bg} $ and $ {\mathcal L}_{fg} $ prevents the model from mis-activating class-specific background snippets that own similar static content to action snippets and promotes its ability for action modeling, simultaneously.  Here, we propose a background gradient enhancement strategy applied in $ {\mathcal L}_{bg} $ to achieve the adversarial learning procedure. Similar to GRL~\cite{G4_unsupervised}, the designed strategy aims to modify the gradient in the backward procedure. 
 To clarify our strategy, we first formulate the generalized gradient updating process:
\begin{equation}
w_{new}=w_{old}-\eta\cdot\frac{\partial {\mathcal L}}{\partial w_{old} },
\label{eq7}
\end{equation}
where $ w $ represents a learnable parameter, $ \eta $ is updating rate, and $ {\mathcal L} $ means optimization loss function. As mentioned in section 3.1, both the classifier and TAM are utilized to predict the action probability of each snippet. Naturally, the proposed background gradient enhancement strategy can be implemented from two perspectives, i.e. 1) modification of the weights of the classifier, and 2) revision of the parameters of TAM.  

\textbf{Modification of the classifier’s weights.} Intuitively, we can adopt an additional background classification loss that optimizes the model by setting each snippet as background, enhancing the background gradient of the classifier. And the loss is formulated as: 
\begin{equation}
{\mathcal L}_{bg}^v =-\sum_{c=1}^{C+1}\tilde{y}_c^{bg} \cdot \mathrm {log}(\uptau(\frac{1}{T}\sum_{i=1}^{T}y_i)),
\label{eq_7}
\end{equation}
where $ \tilde{y}_c^{bg}  $ is one-hot label of the background. With the background loss $ {\mathcal L}_{bg}^v $, the background gradient of classifier for each snippet is raising during the backward procedure (calculated by Eq.~\ref{eq7}), increasing the difficulty of the model to recognize the action content.

\textbf{Revision of the weights of TAM.} TAM is utilized to actively predict the score that each snippet belongs to action or background, so the revision of the parameters of TAM can effectively achieve the background gradient enhancement strategy. Distinct from the manner adopted on the classifier, the proposed strategy based on the modification of the weight of TAM cannot be directly realized by introducing an auxiliary loss.  For achieving the goal, we first analyze the gradient backward process of background classification loss $ {\mathcal L}_{bg} $, and then implement the background gradient enhancement strategy based on the analysis. 

(1) \textit{Analysis}. Take the attention weight $ w_{att}^r $ in RGB modality as an example, we formulate the derivation result with $P_{bg}^{C+1}$ and $y_{r,i}^{C+1}$ as:
\begin{equation}
\frac{\partial \mathcal L_{bg}}{\partial w_{att}^r}=\sum_{i=1}^T\{\frac{(1-P_{bg}^{C+1})}{-2}\cdot\frac{(y_{r,bg}^{C+1}-y_{r,i}^{C+1})}{N_b}\cdot a_{r}^i(1-a_{r}^i)\cdot X_{e,i}^r\}=\sum_{i=1}^T{g_i},
\label{eq_8}
\end{equation}
where $y_{r,bg}^{C+1}=\frac{1}{N_b} \sum_{i=1}^T(1-a_{r}^i)\cdot y_{r,i}^{C+1}$ is the video-level background score, $X_{e,i}^r$ and $ g_i $ represent the embedded feature and the gradient of the $ i $-th snippet, respectively. The Eq.~\ref{eq_8} indicates that $\frac{\partial E}{\partial w_{att}^r}$ is accumulated with the gradient of each snippet, and the gradient direction of $ g_i $ from the $ i $-th snippet depends on the difference $y_{r,bg}^{C+1}-y_{r,i}^{C+1}$. Concretely, if the background score $ y_{r,i}^{C+1} $ of the $ i $-th snippet is higher than the video-level one, i.e. $(y_{r,bg}^{C+1}-y_{r,i}^{C+1})<0$, the gradient  $ g_i $ is calculated as a positive by Eq.~\ref{eq_8}. Correspondingly, the temporal attention weight $ w_{att}^r $ calculated by Eq.~\ref{eq7} is lower than it was before, leading to the decrease of foreground score $ a_r^i $ and the rise of the background score $ (1-a_r^i )$ calculated in Eq.~\ref{eq_3}.

(2) \textit{Implementation}. Based on the above analysis, we attempt to change the difference $(y_{r,bg}^{C+1}-y_{r,i}^{C+1})$ to $(-y_{r,bg}^{C+1}-y_{r,i}^{C+1})$ in Eq.~\ref{eq_8}, for reducing the gradient of temporal attention weight $ w_{att}^r $, which raises the background score $ 1-a_r^i $. For clarity, we formulate the weight updating procedure before and after the gradient transformation of the $ i $-th snippet as follows:
\begin{equation}
\begin{aligned}
w_{new}&=w_{old}+\eta\cdot\frac{val\cdot(y_{r,bg}^{C+1}-y_{r,i}^{C+1})}{ 2 },\\
\rightarrow  w_{new}^{'}&=w_{old}^{'}+\eta\cdot\frac{val\cdot(-y_{r,bg}^{C+1}-y_{r,i}^{C+1})}{ 2 },
\end{aligned}
\label{eq_9}
\end{equation}
where $ val>0 $ represents the remaining terms in Eq.~\ref{eq_8}, and the video-level background score is positive. From Eq.~\ref{eq_9}, we can learn that $ w_{new}>w_{new}^{'} $. And the manner, modifying $(y_{r,bg}^{C+1}-y_{r,i}^{C+1})$, increases the gradient that tends to background by $ \eta\cdot val\cdot y_{r,bg}^{C+1}$ for the $ i $-th snippet. Fortunately, with a clever construction procedure, the background gradient enhancement strategy represented as Eq.~\ref{eq_9} can be approximated by setting the normalized factor $ N_b $ as $ N_f=\sum_{i=1}^Ta_i $ in Eq.~\ref{eq_3}, and the new gradient formula is defined as:
\begin{equation}
\frac{\partial \mathcal L_{bg}}{\partial w_{att}^r}=\sum_{i=1}^T\{\frac{(1-P_{bg}^{C+1})}{-2}\cdot\frac{(-y_{r,bg}^{C+1}-y_{r,i}^{C+1})}{N_b}\cdot a_{r}^i(1-a_{r}^i)\cdot X_{e,i}^r\},
\label{eq_10}
\end{equation}
where $y_{r,bg}^{C+1}$ is modified as $\frac{1}{N_f}\sum_{i=1}^T(1-a_{r}^{i})\cdot y_{r,i}^{C+1}$, and the normalized factor of $ P_{bg} $ is also set as $ N_f $. Notably, we solely revise the normalized factor $ N_b $ in Eq.~\ref{eq_3}, and the distribution tendency of the background probability is retained.  As a result, this is an ingenious way to realize the proposed strategy, leading to the adversarial learning procedure for the model. Here, we adopt the strategy implemented by modifying the weights of TAM in our method, and the extensive experiments demonstrate its effectiveness. 

\noindent\textbf{Discussion:} Notably, GRL~\cite{G4_unsupervised} is applied to map the features of the target and the source domains into a similar distribution space by reversing the gradient. And our work is aiming at improving the model performance to enlarge the discrepancy between action and background, so GRL is not suitable for our task. Functionally, GRL solely revises the difference $(y_{r,bg}^{C+1}-y_{r,i}^{C+1})$ to $(y_{r,i}^{C+1}-y_{r,bg}^{C+1})$  in Eq.~\ref{eq_8}. In this condition, if the $i$-th snippet owns highly background score $y_{r,i}^{C+1}$, the gradient direction of $ g_i $ would be toward to the foreground instead of the background, and vice versa. Consequently, the excessive confusion created by GRL blocks the model from learning action information. Distinct from the GRL, the proposed strategy modifies the difference $(y_{r,bg}^{C+1}-y_{r,i}^{C+1})$  to  $(-y_{r,bg}^{C+1}-y_{r,i}^{C+1})$, which  impartially increases the background gradient of each snippet by the same magnitude, i.e. $ y_{r,bg}^{C+1} $. In this way, the background loss  $ {\mathcal L}_{bg}  $ confuses the model by forcing the gradient of each snippet to tend to be background in backward procedure.  Under this condition, the model, driven by the foreground loss $ {\mathcal L}_{fg}  $, has to improve the ability to focus on action information to mitigate the interference of background-gradient enhancement.

\subsection{Action localization}
In inference stage, we solely run a forward procedure with the base branch to get a set of action proposals for untrimmed videos in test dataset. From the base branch, we obtain the class-activate scores  $y \in \mathbb R^{T \times (C+1)}$ and temporal attentions $a \in \mathbb R^T$.  And a SoftMax operation is perfomed on $y$ to get the normalized action probability $\bar y$. According to video-level label $ c $ (over a threshold $ \rho_{cls} $) predicted by BB, we capture the class-specific score $ \bar y^{c}  \in \mathbb R^T$. Following the previous work~\cite{w18_em}, we calculate the final localization scores $ S_l $ by fusing$\bar y^{c}$ and $ a $:
\begin{equation}
S_l=\varepsilon\cdot\bar y^{c}+(1-\varepsilon)\cdot a,
\label{eq18}
\end{equation}
where $\varepsilon$ is a hyper-parameter. Next, we apply multi-thresholds on $S_l$ to generate the set $\{s_i,e_i\}$ of each action instance. In addition, Non-Maximum Suppression (NMS) is utilized to filter out some redundancy proposals. For the confidence score $q_i$, we compute it with the general operation mentioned in previous work~\cite{w4_completeness}.

\section{Experiments}

\subsection{Dataset and evaluation }
\textbf{Dataset:} We evaluate the proposed approach in two benchmarks, i.e. THUMOS14~\cite{thumos} and ActivityNet1.2~\cite{activitynet}. As for THUMOS14, it contains 200 videos in the validation set and 213 videos in the test set with 20 categories. Following the previous~\cite{w11_tscn}, we utilized the validation set for training and evaluate our method in the test set. In particular, this dataset is particularly challenging as it includes multiple action instances even with different category labels in a long untrimmed video. ActivityNet1.2 consists of 4819 training videos and 2383 validation videos with 100 categories. By convention~\cite{w14_um}, we train our model with the training set and test on the validation set. In contrast to THUMOS14, it is a large-scale benchmark and generally contains an action instance in a video.

\textbf{Evaluation Metrics:} We follow the standard evaluation protocol that reports the performance of our method with the Mean Average Precision (mAP) values under various intersection over union (IoU) thresholds. And we utilize the evaluation code provided by ActivityNet to validate the performance in ActivityNet1.2.

\subsection{Implementation details}
We adopt the I3D~\cite{i3dandk} network as our feature extractor, which is pre-trained on Kinetics. For a fair comparison, we have not fine-tuned the I3D network on THUMOS14 and ActivityNet1.2. We only fuse their corresponding CAS and temporal attention for training.  Adam optimizer is adopted to optimize the networks with a weight decay of 1e-3. The learning rate is initialized as 1e-3 and decreased to 1e-4 after half of the training iterations. For hyperparameters, we set $\varepsilon$, $ \beta $, and $ k $ as 0.5, 0.1, and 4, respectively. And $\lambda$ is set as 0.1 and 0.02 for THUMOS14  and ActivityNet1.2, respectively. During inference, the thresholds of the class probability and NMS are set as 0.1 and 0.5 respectively.  All the experiments are conducted on Pytorch framework with one 3090 GPU.
 
\begin{table*}[!t]
	\renewcommand{\arraystretch}{1}
	\caption{Localization performance comparison with state-of-the-art methods on the THUMOS14 test set.}
	\label{table_1}
	\setlength\tabcolsep{1mm}
	\centering
	\begin{tabular}{c|c|c c c c c c c|c|c|c}
		\hline
		\hline
		\multirow{2}*{\textbf{Supervision}\centering}&\multirow{2}{3cm}{\textbf{Method}\centering} & \multicolumn{7}{c|}{\textbf{mAP(\%)@IoU }} & \multicolumn{3}{c}{\textbf{AVG mAP(\%)}}\\ 
		\cline{3-12} 	       	
		& & \textbf{\emph{0.1}} & \textbf{\emph{0.2}} & \textbf{\emph{0.3}} & \textbf{\emph{0.4}} & \textbf{\emph{0.5}} &\textbf{ \emph{0.6}} & \textbf{\emph{0.7}}   & \textbf{\emph{(0.1:0.5)}}& \textbf{\emph{(0.3:0.7)}} & \textbf{\emph{(0.1:0.7)}}\\
		\hline
		\hline
		\multirow{3}*{Fully\centering}
		& SSN (2017) \cite{full1_ssn}   & 60.3 & 56.2 & 50.6 & 40.8 & 29.1 & - & - & 47.0 & 28.7 & 37.8\\
		& BSN (2018) \cite{full2_bsn}  & - & - & 53.5 & 45.0 & 36.9 & 28.4 & 20.0 & - & 38.5 & -\\
		& P-GCN (2019) \cite{full10_graph}  & 69.5 & 67.5 & 63.6 & 57.8 & 49.1 & - & - & 61.5 & - & -\\
		\hline
		\multirow{11}*{Weakly\centering}
		
		& TSCN (2020) \cite{w11_tscn} & 63.4 & 57.6 & 47.8 & 37.7 & 28.7 & 19.4 & 10.2 & 47.0 & 28.7 & 37.8\\
		
		& UM (2021) \cite{w14_um}  & - & - & 46.9 & 39.2 & 30.7 & 20.8 & 12.5 & - & 30.0& - \\
		
		
		&ACSNet (2021) \cite{w1_acsnet}      &  -    & -    & 51.4 & 42.7 & 32.4 & 22.0  & 11.7 & - & 32.0 & -\\
		
		&AUMN (2021) \cite{w10_aum}   & 66.2 & 61.9 & 54.9 & 44.4 & 33.3 & 20.5 & 9.0 & 52.1 & 32.4 & 41.4\\
		
		&CoLA (2021) \cite{w2_cola} & 66.2 & 59.5 & 51.5 & 41.9 & 32.2 & 22.0 & 13.1 & 50.3 & 32.1& 40.9 \\
		
		&D2-Net (2021) \cite{w7_d2} & 65.7 & 60.2 & 52.3 & 43.4 & 36.0 & - & - & 51.5 & -& -\\
		& FAC-Net (2021) \cite{w6_foreground} & 67.6 & 62.1 & 52.6 & 44.3 & 33.4 & 22.5 & 12.7 & 52.0 & 33.1& 42.2 \\
		& CO2-Net (2021) \cite{CO2} & \textbf{70.1} & 63.6 & 54.5 & 45.7 & 38.3 & 26.4 & 13.4 & 54.4 & 35.6& 44.6 \\
		& ACGNet (2022) \cite{acgnet} & 68.1 & 62.6 & 53.1 & 44.6 & 35.7 & 22.6 & 12.0 & 52.6 & 33.4& 42.5 \\
		& FTCL (2022) \cite{FTCL} & 69.6 & 63.4 & 55.2 & 45.2 & 35.6 & 23.7 & 12.2 & 53.8 & 34.4& 43.6 \\
		& \textbf{Ours} & 69.7 & \textbf{64.5} & \textbf{58.1} & \textbf{49.9} &  \textbf{39.6} &  \textbf{27.3} & \textbf{14.2 }&\textbf{56.3} &\textbf{37.8} & \textbf{46.1} \\
		\hline
		\hline
		
	\end{tabular}
\end{table*}

\subsection{Comparison with state-of-the-arts}
We compare the proposed approach with recent weakly supervised state-of-the-art methods and several fully supervised. And the results are reported in Tab.~\ref{table_1} and Tab.~\ref{table_2}. It can be seen that our method outperforms the mentioned weakly supervised approaches in terms of almost all metrics. Specifically,  our method leads to 5.4\% average mAP (0.3:0.7) improvement over AUMN~\cite{w10_aum}, which is also designed to model the action of the video. Moreover, we improve the 1.9\% average mAP (0.1:0.5) and 2.2\%average mAP (0.3:0.7), compared with CO2-Net~\cite{CO2}, in which the late fusion way is similar to ours. Compared with several fully supervised counterparts, our method still obtains competitive performance at low temporal IoU. Consequently, the above results demonstrate the effectiveness of the proposed method. ActivityNet1.2 is a large dataset, we also compare several recent works in Tab.~\ref{table_2}. It can be seen that our method still achieves great performance at low IoU (0.5). However, our method does not achieve the same improvement on ActivityNet1.2 as it does on THUMOS14. We argue the reasons are that the background snippets on ActivityNet1.2 are easier to distinguish compared to THUMOS14 and the annotations of ActivitNet1.2 are coarser than those of THUMOS14~\cite{CO2}. Notably, the proposed method mainly contributes to the mitigation of background disturbance. As a result, our method effectively finds out the action instance (i.e. high performance at low IoU) but can not achieve better localization completeness similar to the performance on THUMOS14.

\begin{table}[htbp]
	\renewcommand{\arraystretch}{1}
	\setlength{\tabcolsep}{3.2mm}
	\caption{Comparison of the proposed method with other approaches on the ActivityNet1.2 validation set. AVG indicates average mAP from IoU 0.5 to 0.95 with a 0.05 increment.}
	\label{table_2}
	\setlength\tabcolsep{1mm}
	\centering
	\begin{tabular}{c|c|c c c |c}
		\hline
		\hline
		\multirow{2}*{\textbf{Supervision}\centering}&\multirow{2}{2.8cm}{\textbf{Method}\centering} & \multicolumn{4}{c}{\textbf{mAP@IoU}}\\ 
		\cline{3-6} 	       	
		& & \textbf{\emph{0.5}} &\textbf{\emph{0.75}} & \textbf{\emph{0.95}}& \textbf{\emph{AVG}}\\
		\hline
		\hline
		\multirow{1}*{Fully\centering}&SSN (2017) \cite{full1_ssn} &41.3 &27.0 &6.1 &26.6\\
		\hline
		\multirow{7}*{Weakly\centering}&DGAM (2020) \cite{w3_gam} &41.0 &23.5 &5.3 &24.4\\
		&TSCN (2020) \cite{w11_tscn} &37.6 &23.7 &5.7 &23.6\\
		&BS (2020) \cite{w17_backgroundsuppression} &38.5 &24.2 &5.6 &24.3\\
		&HAM-Net (2021) \cite{ham} &41.0 &24.8 &5.3 &25.1\\
		&UM (2021) \cite{w14_um} &41.2 &25.6 &6.0 &25.9\\
		&ASCNett (2021) \cite{w1_acsnet} &40.1 &\textbf{26.1} &\textbf{6.8} &\textbf{26.0}\\
		&\textbf{Ours} & \textbf{41.6} & 24.8 & 5.4  & 25.2 \\
		\hline
		\hline
	\end{tabular}
\end{table}

\subsection{Ablation study}
By convention~\cite{w10_aum}, to look deeper into the proposed method, we conduct the corresponding ablation studies on THUMOS14. 
\begin{table}[htbp]
	\renewcommand{\arraystretch}{1}
	\caption{Evaluation of the effectiveness of the designed component under multiple IoU thresholds from 0.3 to 0.7 with an increase of 0.1.}
	\label{table_3}
	\setlength\tabcolsep{1mm}
	\centering
	\begin{tabular}{c|c c c c c |c}
		\hline
		\hline
		\multirow{2}{2.5cm}{\textbf{Method}\centering} & \multicolumn{6}{c}{\textbf{mAP@IoU}}\\ 
		\cline{2-7} 	       	
		& \textbf{\emph{0.3}}&\textbf{\emph{0.4}} & \textbf{\emph{0.5}} &\textbf{\emph{0.6}} & \textbf{\emph{0.7}}& \textbf{\emph{AVG(0.3:0.7)}}\\
		\hline
		\hline
		BL &51.4 &42.1 &31.3 &20.6 &10.5 &31.2 \\
		BL+BGES &56.1 &45.5 &34.2 &24.2 &12.9 &34.5 \\
		TEN &53.7 &46.1 &35.6 &24.8 &13.2 &34.7\\
		TEN+BGES &\textbf{58.1} &\textbf{49.9} &\textbf{39.6} &\textbf{27.3} &\textbf{14.2} &\textbf{37.8}\\
		\hline
		\hline
	\end{tabular}
\end{table}

\begin{table}[htbp]
	\renewcommand{\arraystretch}{1}
	\caption{Comparisons of different types of combinations for the gradient modification strategy and the designed component.}
	\label{table_4}
	\setlength\tabcolsep{1mm}
	\centering
	\begin{tabular}{c|c c c c c |c}
		\hline
		\hline
		\multirow{2}{2.5cm}{\textbf{Method}\centering} & \multicolumn{6}{c}{\textbf{mAP@IoU}}\\ 
		\cline{2-7} 	       	
		& \textbf{\emph{0.3}}&\textbf{\emph{0.4}} & \textbf{\emph{0.5}} &\textbf{\emph{0.6}} & \textbf{\emph{0.7}}& \textbf{\emph{AVG(0.3:0.7)}}\\
		\hline
		\hline
		BL &51.4 &42.1 &31.3 &20.6 &10.5 &31.2 \\
		TEN &53.7 &46.1 &35.6 &24.8 &13.2 &34.7\\
		\hline
		BL+GRL &53.2 &43.6 &33.2 &21.8 &10.9 &32.5 \\
		TEN+GRL &55.7 &46.0 &35.2 &22.6 &12.1 &34.3\\
		\hline
		BL+BVL &54.7 &46.1 &35.1 &22.5 &11.3 &34.0 \\
		TEN+BVL &55.9 &47.1 &35.6 &24.5 &12.8 &35.2\\
		TEN+BVL+BGES &\textbf{57.4} &\textbf{49.5} &\textbf{39.0} &\textbf{27.1} &\textbf{14.1} &\textbf{37.5}\\
		\hline
		\hline
	\end{tabular}
\end{table}

\begin{table}[htbp]
	\renewcommand{\arraystretch}{1}
	\caption{Ablation studies of different kinds of object loss on attention (ATT) and class activate scores (CAS).}
	\label{table_5}
	\setlength\tabcolsep{1mm}
	\centering
	\begin{tabular}{c|c| c c c c c c c |c}
		\hline  
		\hline
		\multirow{2}*{\textbf{Object}\centering} &\multirow{2}{0.9cm}{\textbf{Loss}\centering} & \multicolumn{8}{c}{\textbf{mAP@IoU}}\\ 
		\cline{3-10} 	       	
		& &\textbf{\emph{0.1}}&\textbf{\emph{0.2}}& \textbf{\emph{0.3}}&\textbf{\emph{0.4}} & \textbf{\emph{0.5}} &\textbf{\emph{0.6}} & \textbf{\emph{0.7}}& \textbf{\emph{0.1:0.7}}\\
		
		\hline
		\hline
		\multirow{2}*{ATT\centering} & MAE & 68.1 &64.0 &56.5 &46.5 &35.3 &24.5&13.2 & \textbf{44.0}\\
		& MSE & 67.8 &63.9 &56.2 &46.4 &35.0 &24.4&13.2 &  43.8\\
		\hline
		\multirow{3}*{CAS\centering} & MAE & 67.9&63.7 &56.4 &46.4 &35.2 &24.7&13.3 & 43.9\\
		& MSE & 68.0 &64.0 &56.3 &46.5 &35.1 &24.5&13.3 & 43.9\\
		& KL & \textbf{69.4}&\textbf{64.0} & \textbf{56.9} &\textbf{48.6} & \textbf{38.1} &\textbf{26.8}& \textbf{14.3} &  \textbf{45.5}\\
		\hline
		\hline
	\end{tabular}
\end{table}

\begin{table}[htbp]
	\renewcommand{\arraystretch}{1}
	\caption{Evaluation of different numbers of the sampling interval $ k $ in TEN in terms of average mAP under IoU thresholds (0.1:0.1:0.7).}
	\label{table_6}
	\setlength\tabcolsep{1mm}
	\centering
	\begin{tabular}{c|c c c c c c c |c |c}
		\hline  
		\hline
		\multirow{2}{0.5cm}{$ k $\centering} & \multicolumn{7}{c|}{\textbf{mAP@IoU}}&\multicolumn{2}{c}{\textbf{AVG mAP(\%)}}\\ 
		\cline{2-10} 	       	
		&\textbf{\emph{0.1}}&\textbf{\emph{0.2}}& \textbf{\emph{0.3}}&\textbf{\emph{0.4}} & \textbf{\emph{0.5}} &\textbf{\emph{0.6}} & \textbf{\emph{0.7}}& \textbf{\emph{0.1:0.5}}& \textbf{\emph{0.3:0.7}}\\
		
		\hline
		\hline
		2 & \textbf{70.0} &\textbf{ 65.4}&\textbf{58.5} &48.1 &36.7 &26.1 &13.8&55.7 &36.6\\
		3 & 69.2 & 64.0&57.3 &49.1 &38.8 &27.0 &\textbf{14.5}&55.7 & 37.4\\
		4 & 69.7 & 64.5&58.1 &\textbf{49.9} &\textbf{39.6} &\textbf{27.3} &14.2&\textbf{56.3} &\textbf{37.8}\\
		5 & 68.5 & 63.7&57.7 &49.4 &39.4 &27.1 &13.9&55.7 &37.5\\
		\hline
		\hline
	\end{tabular}
\end{table}

\begin{figure}[t]
	\centering
	\includegraphics[scale=0.28]{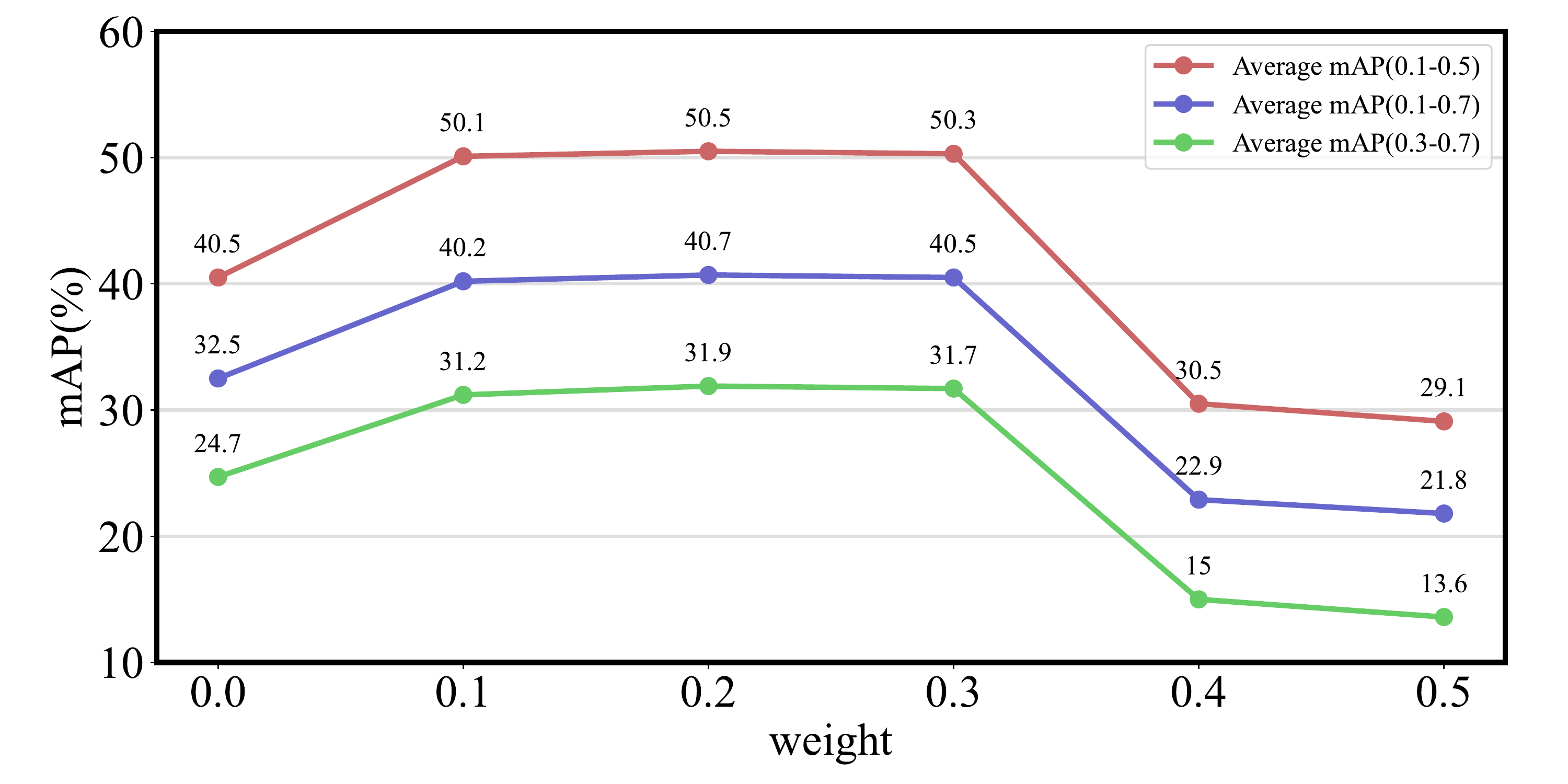}
	\caption{Ablation results of the baseline with different background classification loss weights on the THUMOS14 dataset. }
	\label{visual_0}
\end{figure}
\textbf{Effectiveness of each component.} To verify the effect of each design mentioned in the method section, we conduct corresponding ablation studies and report the results in Tab.~\ref{table_3}. Here, we treat “BL” as the baseline of our method and set “BGES” as the background gradient strategy based on the modification of attention mechanism weights, and “TEN” represents the temporal enhancement network. From Tab.~\ref{table_3}, we can find that BGES obtains a 3.3\% improvement over the baseline, and TEN also surpasses the baseline by 3.5\%. BGES mainly promotes the performance on the lower IoU (mAP@IoU 0.3), while TEN improves the effectiveness on the higher one (mAP@IoU 0.5:0.7). And the full mode by combining BGES and TEN largely surpasses the baseline by 6.6\%, and it outperforms both variants, proving that the above two components are not redundant. 

\textbf{Effect of different manners for gradient modification.} We compare the different gradient-modification ways mentioned in the approach section, and the related results are reported in Tab.~\ref{table_4}. Note that BVL is the proposed strategy shown in Eq.~\ref{eq_7}. We first apply GRL to the base model, and it outperforms the baseline on the low IoU but is not significant. Furthermore, we assemble GRL to TEN, and this manner solely improves the performance on low IoU but worsens the results on the average mAP@IoU (0.3:0.7), which indicates GRL solely facilitates the model to focus on the most discriminative action snippets but ignore the rest action ones. As for BVL, we introduce it to the base model and observe an increase (2.8\% on the average mAP@IoU (0.3:0.7)) compared with the baseline. We also apply BVL to TEN for further comparison, and it surpasses BVL and TEN by 1.2\% and 0.5\% on mAP@IoU (0.3:0.7), respectively. Notably, the improvement of the manner that applies BVL to TEN is not significant. For this, the reason may lie in the weight-updating conflict between BVL and the KL loss in the backward procedure. Moreover, we plug both the strategy BVL and BGES into TEN, but this manner still does not outperform the result reported in the last row of Tab.
~\ref{table_3}. As a result, we solely adopt the strategy implemented by modifying the weights of TAM.

\textbf{The influence of the weight $ \lambda $ of background classification loss.} To better verify the effectiveness of BGES, we also investigate the influence of $ \lambda $ by setting it from 0.0 to 0.5 in the baseline method. The results are shown in Fig.~\ref{visual_0}. It can be seen that the performance of the network has dropped obviously when the value is larger than 0.3. Therefore, it is evidence that simply increasing the value of the weight $\lambda$ can not achieve the adversarial learning procedure similar to BGES for improving the performance of action localization.

\textbf{The effectiveness of MAE, MSE, and KL.} We adopt different losses on temporal attention (ATT) and class activate scores (CAS) in Tab.~\ref{table_5}. It can be seen that all of the constraint losses on ATT or CAS improve the performance of action localization. As for the CAS, KL loss outperforms MAE and MSE by 1.6\% on the average mAP (0.1:0.7), which indicates that KL is beneficial to mutual learning among the branches. As for the ATT, MAE loss shows a better performance than MSE loss. 

\textbf{Ablations on different sampling intervals $ k $.}We set the $ k $ from 2 to 5 and the results are shown in Tab.~\ref{table_6}.  It can be seen that our methods consistently achieve the best performance on average mAP (0.1:0.5) and mAP (0.3:0.7) when $ k $ =4. 

\begin{figure}[t]
	\centering
	\includegraphics[scale=0.25]{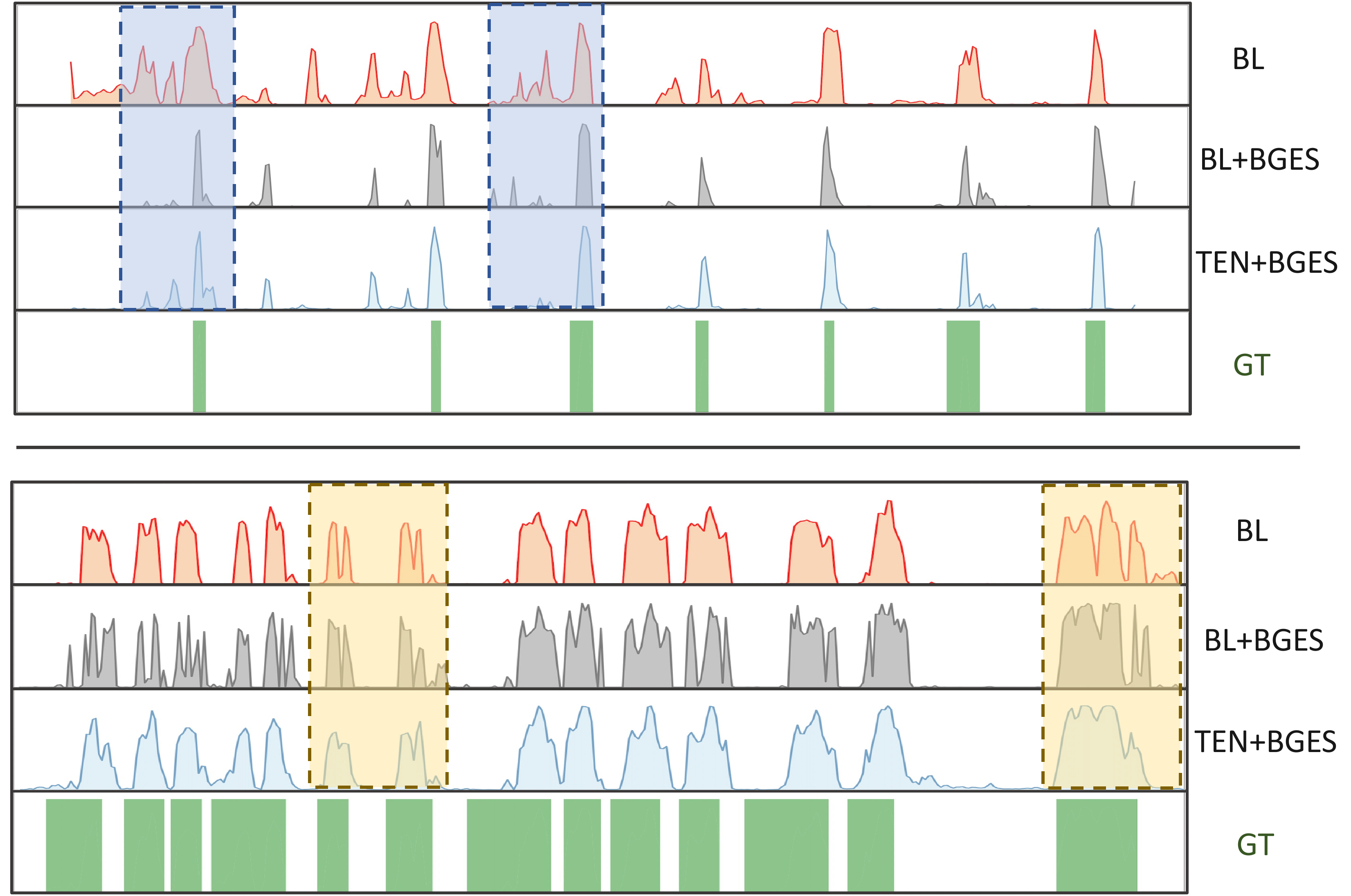}
	\caption{The illustration of qualitative results of different types for the combination of the designed components on two instances of soccer-penalty and clean-and-jerk. And the blue and the yellow rectangles show the different localization results of various methods.}
	\label{visual_1}
\end{figure}

\subsection{Qualitative results}
To better demonstrate the effectiveness of the proposed method, we visualize several examples of localized action instances in Fig.~\ref{visual_1}. For the first instance of the soccer penalty, compared with the baseline, the proposed strategy BGES and the full method by combining BGES and TEN can effectively alleviate the interference of the background (shown in blue rectangles). In the second example of clean-and-jerk, compared with the baseline and BGES, the proposed TEN successfully construct the temporal relation of affinity snippets and make the detected results more smooth. Consequently, all of the visual results verify the effect of the proposed method.

\section{Conclusion}
In this work, we explore an adversarial learning strategy for WTAL and propose a novel temporal enhancement network (TEN) based on the background gradient enhancement strategy (BGES). We utilized the background gradient enhancement strategy to achieve the adversarial learning procedure, aiming to relieve the background disturbance in the model learning process. Additionally, the temporal enhance network is designed to construct the temporal consistency of continuous snippets, improving the ability to localize complete action. Finally, we conduct extensive experiments to verify the effectiveness of the proposed method.

\section{ACKNOWLEDGEMENTS}
The work described in this paper was partially supported by the National Natural Science Foundation of China (Grant no.62176031) and the Fundamental Research Funds for the Central Universities (Grant no. 2021CDJQY-018).

\bibliographystyle{ACM-Reference-Format}
\bibliography{acmart.bib}
\end{document}